\documentclass[lettersize,journal]{IEEEtran}
\usepackage{amsmath,amssymb,amsfonts}
\usepackage{algorithmic}
\usepackage{array}
\usepackage{subfigure}

\usepackage{graphicx}
\usepackage{epstopdf}
\usepackage{epsfig}
\usepackage{caption}
\usepackage{textcomp}
\usepackage{stfloats}
\usepackage{booktabs}
\usepackage{xcolor}
\usepackage{cuted}

\def\BibTeX{{\rm B\kern-.05em{\sc i\kern-.025em b}\kern-.08em
    T\kern-.1667em\lower.7ex\hbox{E}\kern-.125emX}}

\begin{document}
\title{Study of Enhanced MISC-Based Sparse Arrays with High uDOFs and Low Mutual Coupling}
\author{Xueli Sheng, Dian Lu, Yingsong Li, \emph{Senior Member, IEEE}, Rodrigo C. de Lamare, \emph{Senior Member, IEEE}
\thanks{The work was supported in part by the National Key R\&D Program of China under Grant 2022YFC2807804, and in part by the National Natural Science Foundation of China under Grant U20A20329. {\it (Corresponding author: Yingsong Li.)} }
%\thanks{Xueli Sheng and Dian Lu, are both with 1) the National Key Laboratory of Underwater Acoustic Technology, Harbin Engineering University, Harbin 150001, China; 2) the Key Laboratory for Polar Acoustics and Application of Ministry of Education (Harbin Engineering University), Ministry of Education, Harbin 150001, China; and 3) the College of Underwater Acoustic Engineering, Harbin Engineering University, Harbin 150001, China (e-mail: shengxueli@hrbeu.edu.cn; ludian@hrbeu.edu.cn).}
\thanks{Xueli Sheng and Dian Lu, are both with the College of Underwater Acoustic Engineering, Harbin Engineering University, Harbin 150001, China (e-mail: shengxueli@hrbeu.edu.cn; ludian@hrbeu.edu.cn).}
\thanks{Yingsong Li is with the School of Electronic and Information Engineering, Anhui University, Hefei 230601, China (e-mail: liyingsong@ieee.org).}
\thanks{Rodrigo. C. de Lamare is with the Centre for Telecommunications Research (CETUC), Pontifical Catholic University of Rio de Janeiro (PUC-Rio), G$\mathbf{\acute{a}}$vea, 22451-900, Brazil, and the Department of Electronic Engineering, University of York, York, YO10 5DD, UK (e-mail: delamare@cetuc.puc-rio.br).}
}

%\markboth{IEEE Communication Letters,~Vol.~, No.~, June~2023}.
%\fancyhead[RO]{IEEE Communication Letters,~Vol.~, No.~, June~2023}.
%\fancyhead[LE]{SHENG et al.: AN ENHANCED MISC-BASED SPARSE ARRAY WITH HIGH uDOFs AND LOW MUTUAL COUPLING}.

\maketitle
\begin{abstract}
In this letter, inspired by the maximum inter-element spacing (IES) constraint (MISC) criterion, an enhanced MISC-based (EMISC) sparse array (SA) with high uniform degrees-of-freedom (uDOFs) and low mutual-coupling (MC) is proposed, analyzed and discussed in detail. For the EMISC SA, an IES set is first determined by the maximum IES and number of elements. Then, the EMISC SA is composed of seven uniform linear sub-arrays (ULSAs) derived from an IES set. {An} analysis of the uDOFs and weight function shows that, the proposed EMISC SA outperforms the IMISC SA in terms of uDOF and MC. Simulation results show a significant advantage of the EMISC SA over {other existing SAs}.
\end{abstract}


\section{Introduction}
\IEEEPARstart{A}{rray} signal processing techniques are gaining popularity in various applications, including radar \cite{1}, sonar \cite{2,3} etc. Traditional uniform linear arrays (ULAs) are usually utilized for beamforming [1],\cite{jioccm,ccg,smcg,jiostap,smbeam,wcccm,wlmwf,wljio,l1cg,rdcapon,wlccm,spstap,locsme,okspme,lrcc} and DOA estimation [4]. However, limited by the Rayleigh criterion, a $K$-element ULA can only resolve at most N-1 sources \cite{4,5}. {$N$-1 sources} can provide sufficient uniform degrees of freedom (uDOFs), allowing fewer sensors to achieve the spatial resolution of more targets \cite{6}. Consequently, with the DCA concept applied to the existing SAs such as nested arrays (NAs) \cite{NESTED,8} and coprime arrays (CAs) \cite{ePCA,ICNA}, NAs can generate $O(\frac{K^2}{2})$ uDOFs for a $K$-element ULA but are vulnerable to {mutual coupling (MC)}. In contrast, CAs have {significantly fewer} uDOFs than {NAs} but are more resistant to the MC effect.

Numerous SAs have been developed [11]-[14] through the prototype NAs and CAs. The early SAs are mostly aimed at how to increase the uDOFs. {For few sensors, the minimum redundancy arrays (MRAs) is the most favorable choice due to the maximum uDOFs}. NAs such as the improved NA (INA) [11], the CA having displaced-subarrays (CADiS) [12], the augmented NAs (ANA) [13], {the two side-extended NA (TSENA) [14], and the flexible extended NA with multiple subarrays (f-ENAMS)[15]} all demonstrate excellent enhancement on uDOFs. Nevertheless, {the MC effect between the elements of small separation should not be neglected}. Consequently, various SAs have been designed to reduce MC, such as the super NAs (SNAs) [15]-[18], the extended padded CAs (ePCAs) [19], and the improved coprime NA (ICNA) [20]. Moreover, another SA design method - the ULA fitting scheme composed of four base layers (UF-4BL) has been suggested with modest MC [21], [22]. {Except for the conventional SA designs}, an SA design based on the MISC criterion is presented and investigated [23]. The MISC-based SAs are developed via an IES set, jointly determined by the maximum IES and number of elements. For uDOFs, an improved MISC (IMISC) SA [24] is proposed with a much higher uDOF.

In this letter, compared with the IMISC SA, an enhanced MISC-based (EMISC) SA that further increases the uDOFs and reduces the MC effect is proposed. Then, relevant analysis is carried out from aspects of uDOF, MC and DOA estimation. Simulation results reflect a great advantage of the EMISC SA over other existing SAs.

Notation: Throughout {this} letter, lower-case (upper-case) characters represent vectors (matrices). Particularly, ${\rm{vec}}({\mathbf{I}}_K)$ is the $K \times K$ identity matrix. The superscripts: $*$, $T$ and $H$ respectively represent the complex conjugate, transpose and conjugate transpose. The operators: $E[\cdot]$ denotes the statistical expectation operator; ${\rm{vec}}(\cdot)$ denotes the vectorization operator; ${\rm{diag}}(\cdot)$ denotes the diagonalization operator; ${\rm{card}}\{\cdot\}$ denotes the element number of a set and $\lfloor \cdot \rfloor$ denotes the floor operator. The symbols: $\odot$ is the Khatri-Rao product; $\lvert\cdot\rvert$ is the absolute value; $\Vert\cdot\Vert_F$ is the Frobenius norm, $\%$ is the remainder and $\cup$ denotes the union operator.

\section{Coarray Concept and Mutual Coupling}\label{fundamental}
\subsection{DCA Model}
First, let us {choose} a $K$-element nonuniform linear array, whose element positions are given by $p_id$, where $p_i \in \mathbb{P}$. $d = {{\lambda}/2}$ denotes the minimum inter-element spacing, with ${\lambda}$ being the wavelength of {the incident} wave.
\begin{equation}
\mathbb{P} = \left\{ {{p_0},p_1, p_2,\dots,p_{K-1}} \right\},
\label{position_set}
\end{equation}

Assume that $N$ far-field, uncorrelated narrowband sources arrive at the array from different bearings $\left\{\theta_i,i=1,\cdots,N\right\}$ corresponding to the powers $\left\{\rho^2_i,i=1,\cdots,N\right\}$.

Then, the received signal can be modeled as
\begin{equation}
{\mathbf{x}} = {\mathbf{A}}{\mathbf{s}} + {\mathbf{n}},
\label{received_data_one_snapshot}
\end{equation}
where $\mathbf{s} \triangleq \left[\mathbf{s}_1,\cdots,\mathbf{s}_N\right]^T$ denotes the incident sources, and ${\mathbf{n}}$ is the additive Gaussian white noise, uncorrelated with the sources.
Besides, $\mathbf{A}\triangleq\left[\mathbf{a}_{\theta_1},\cdots,\mathbf{a}_{\theta_N}\right]$ is the $K \times N$ array manifold matrix, where the corresponding steering vector $\mathbf{a}_{{\theta}_i}=\left[e^{j \frac {2\pi}{\lambda} p_0 d \sin({\theta}_i)},\cdots,e^{j \frac {2\pi}{\lambda} p_{K-1} d \sin({\theta}_i)}\right]^T$, ${\theta}_i$ is the bearing of the $i^{th}$ source.

Next, the covariance matrix (CM) of (\ref{received_data_one_snapshot}) is given by
\begin{equation}
{\mathbf{R}_{\mathbf{x}}} \triangleq E[\mathbf{x}{\mathbf{x}^H}] = \mathbf{A}\mathbf{R}_{\mathbf{s}}{\mathbf{A}^H} + \rho_n^2 \mathbf{I}_K,
\label{covariance_matrix}
\end{equation}
where $\mathbf{R}_{\mathbf{s}}$ is {the CM of the received signal} and $\rho ^2_n$ is the noise power.

Finally, (\ref{covariance_matrix}) is vectorized as (\ref{vectorized_covariance_matrix}), expressed as
\begin{equation}
{\mathbf{y}}\triangleq{\rm{vec}}({\mathbf{R}_{\mathbf{x}}}) = ({\mathbf{A}}^*\odot{\mathbf{A}}){\mathbf{h}}+ \rho_n^2 {\mathbf{1}}_k,
\label{vectorized_covariance_matrix}
\end{equation}
where $\mathbf{1}_k={\rm{vec}}({\mathbf{I}}_K)$, and ${\mathbf{h}} = \left[\rho ^2_1,\cdots,\rho ^2_N\right]^T$ is the received signal of the DCA, defined by (5)~\cite{NESTED}
\begin{equation}
\mathbb{D}= \left\{ {{p_a-p_b},a,b = 0,1, \cdots K-1} \right\}.
\label{DCA}
\end{equation}

\subsection{MC effect}
Considering the MC effect in practical application, (6) is obtained by interpolating the MC matrix $\mathbf{U}$ into (2).
\begin{equation}
{\mathbf{x}} = {\mathbf{U}} {\mathbf{A}}{\mathbf{s}} + {\mathbf{n}},
\label{received_data_one_snapshot_with_coupling}
\end{equation}
where, $\mathbf{U}$ is a $K \times K$ $G$-banded identity matrix [23]-[24].
%%%%%%%%%%%%%%%%%%%%%%%%%%%%%%%%%%%%
\begin{equation}
{\mathbf{U}}_{b,c}=\left\{{\begin{array}{*{20}{l}}
u_{|p_b-p_c|},&|p_b-p_c|\le G,\\
0,&{\rm{elsewhere}},
\end{array}} \right.
\label{c_approximate}
\end{equation}
where, $p_b, p_c \in \mathbb{P}$, $v=|p_b-p_c|\in[0,G]$, and $u_v$ denotes an arbitrary element of $\mathbb{U}$ that is expressed as
%%%%%%%%%%%%%%%%%%%%%%%%%%%%%%%%%%%%
\begin{equation}
\left\{{\begin{array}{*{20}{l}}
u_0=1\textgreater |u_1|\textgreater |u_2|\textgreater\cdots\textgreater|u_G|,\\
|u_i/u_j|=j/i, \quad\hspace{0.3em}\ i,j\in[1,G].
\end{array}}\right.
\label{c_coefficients}
\end{equation}

The MC effect is usually measured by the coupling leakage (CL), given by
%%%%%%%%%%%%%%%%%%%%%%%%%%%%%%%%%%%%
\begin{equation}
M=\frac {||{\mathbf{U}}-{\rm{diag}}({\mathbf{U}})||_F} {||{\mathbf{U}}||_F}.
\label{coupling_leakage}
\end{equation}

The weight function $w(l)$, which outputs the number of element pairs whose IES is \emph{l}, is defined as
\begin{equation}
w(l)={\rm{card}}\{ (p_i,p_j)|p_i-p_j = l;p_i,p_j \in \mathbb{P}\}.
\end{equation}

By the analysis above, there is a close relationship between MC and the weight function. Besides, the first three weight values $w(i), i=1,2,3$ jointly dominate the MC matrix.

\section{Proposed EMISC SA}
Considering that the IMISC SA is composed of six ULSAs [24], another ULSA with an IES of 3 is added {to reduce the MC}. Then, all ULSAs are reasonably rearranged and optimized to enable $w(1)$ equal to one. Thereby, compared with the IMISC SA, we propose an enhanced MISC-based (EMISC) SA, which achieves higher uDOFs and lower MC, and has a closed-form expression of IES set, position set, and uDOF.
\subsection{SA Configuration}
Based on the MISC criterion, {the EMISC SA configuration is schemed by an IES set, which depends on the maximum IES and the number of elements.} Then, the closed-form expression of the IES set $\mathbb{S}_{\text{EMISC}}$ is provided as
\begin{equation}
\mathbb{S}_{\text{EMISC}}=\left\{ \begin{array}{l}
3,\underbrace {2,...,2}_{\frac{L}{4} - 1},\underbrace {L/2+1,...,L/2+1}_{\frac{L}{4} - 1},\\
\underbrace {L,...,L}_{K - L},\underbrace {L/2-1,...,L/2-1}_{\frac{L}{4} - 2},\\
\frac{L}{2} - 2, 1, 3,\underbrace {2,...,2}_{\frac{L}{4} - 2},
\end{array} \right\},
\label{spacing}
\end{equation}
where $K$ is the {number of elements}, $L$ is the maximum IES and
\begin{equation}
L = 4\lfloor \frac{K-4}{6}\rfloor + 4,K\ge10,\\
\label{M}
\end{equation}

According to (\ref{spacing}), the position set is given {by}
\begin{equation}
\begin{aligned}
&\mathbb{P}_{\text{EMISC}} =\\
&\left\{ \begin{array}{l}
\underbrace {0,3}_{{\text{ULA 1, IES=3}}},
\underbrace {5,...,\frac{L}{2} + 1}_{{\text{ULA 2, IES=2}}},
\underbrace {\frac{L}{2} + 3,...,\frac{{{L^2}}}{8} - \frac{L}{4} + 1}_{{\text{ULA 3, IES=}}\frac{L}{2} + 1},\\
\underbrace {\frac{{{L^2}}}{8} + \frac{{L}}{4} + 2,..., - \frac{{7{L^2}}}{8} + (K- \frac{3}{4})L + 2}_{{\text{ULA 4, IES=}}L},\\%444
\underbrace {- \frac{{7{L^2}}}{8} + (K+\frac{1}{4})L + 2,...,- \frac{{3{L^2}}}{4} + (K-1)L + 4 }_{{\text{ULA 5, IES=}}\frac{L}{2}-1},\\%555
\underbrace {- \frac{{3{L^2}}}{4} + (K- \frac{1}{2})L + 2,- \frac{{3{L^2}}}{4} + (K- \frac{1}{2})L + 3}_{{\text{ULA 6, IES=1}}},\\
\underbrace {- \frac{{3{L^2}}}{4} + (K- \frac{1}{2})L + 6,...,- \frac{{3{L^2}}}{4}+KL+2}_{{\text{ULA 7, IES=2}}},\\
\end{array} \right\}
\end{aligned}
\label{Structure}
\end{equation}

It is noted that the EMISC SA {is made up of} seven ULSAs. Fig. 1 represents a specific EMISC SA configuration with an element number $K$ = 10 and the maximum IES $L$ = 8, where sub-\emph{i} ($i=1,...,7$) denotes the \emph{i}-th ULSA, respectively.
%\captionsetup[figure]{labelformat=simple, labelsep=period}
\captionsetup[figure]{name={Fig.}, labelformat=simple, labelsep=period, singlelinecheck=off}
\begin{figure}[!t]
\centering
\includegraphics[width=1\columnwidth,height=1.4cm]{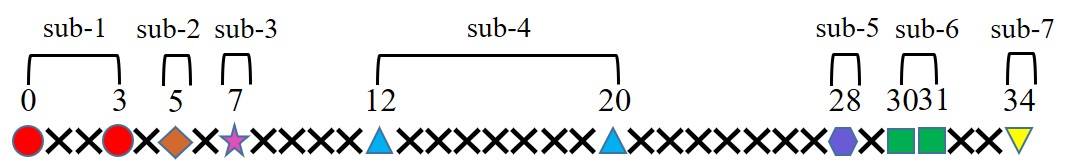}
\caption{The EMISC SA configuration with~$K=10$ and $L=8$.}
\label{EMISC_STRUCTURE}
\end{figure}

\subsection{Consecutive part in the DCA}\label{proof_proposition_SDCA_symmetric}
%%%%%%%%%%%%%%%%%%%%%%%%%%%%%%%%%%%%%%%%%%%%%%%
{Considering the symmetric property of the DCA}, we just need to prove $\mathbb{C}_{\text{EMISC}}^{+} \subset \mathbb{D}_{\text{EMISC}}^{+}$, where $\mathbb{C}_{\text{EMISC}}^{+}$ and $\mathbb{D}_{\text{EMISC}}^{+}$ represent the positive part of $\mathbb{C}_{\text{EMISC}}$ and $\mathbb{D}_{\text{EMISC}}$, respectively. (\ref{ULSAm & ULSAn}) gives the position sets of any two ULSAs.
\begin{equation}
\begin{aligned}
\mathbb{P}_{\text{ULSA}_m} &= \{p_{m}(x),x=1,\dots,K_m\},\\
\mathbb{P}_{\text{ULSA}_n} &= \{p_{n}(x),x=1,\dots,K_n\}.
\label{ULSAm & ULSAn}
\end{aligned}
\end{equation}
where $K_m$ and $K_n$ separately denotes the element number of ULSA-$m$ and ULSA-$n$, and the DCA $\mathbb{D}_{m,n}$ between ULSA-$m$ and ULSA-$n$ is defined as
\begin{equation}
\begin{aligned}
\mathbb{D}_{m,n} = \{p_{n}(y)-p_{m}(x)\, x = 1,\dots,K_m\, y = 1,\dots,K_n\}.
\end{aligned}
\label{D_mn}
\end{equation}
where, $\mathbb{D}_{m,m}=\{p_{m}(x)-p_{m}(1),x=1,\dots,K_m\}$.

Considering the fact that the proposed EMISC SA is composed of seven ULSAs, then, the positive part $\mathbb{D}_{\text{EMISC}}^{+}$ in the DCA $\mathbb{D}_{\text{EMISC}}$ of the proposed EMISC SA is expressed as
\begin{equation}
\begin{aligned}
&\mathbb{D}_{\text{EMISC}}^{+}=\bigcup_{m=1}^{7} \mathbb{D}_{\text{m,n}}^{+},n=m,m+1,\dots,7,\\
\end{aligned}
\label{D_MISC+}
\end{equation}
where $\mathbb{D}_{\text{m,n}}^{+}$ is the positive part of $\mathbb{D}_{m,n}$. $\mathbb{C}_{\text{EMISC}}^{+} \subset \mathbb{D}_{\text{EMISC}}^{+}$ is derived by finding DCAs to make up the consecutive part $[0,- \frac{{3{L^2}}}{4} + (K-\frac{1}{2})L + 3]$.

$\mathbb{D}_{1,1}\cup\mathbb{D}_{1,2}\cup\mathbb{D}_{1,3}\cup\mathbb{D}_{2,3}\cup\mathbb{D}_{5,6}\cup\mathbb{D}_{5,7}\cup\mathbb{D}_{6,6}\cup\mathbb{D}_{6,7}$ generates a consecutive range $\mathbb{R}_1$, as shown in (\ref{cons1}).
\begin{equation}
[0,\frac{L^2}{8}-\frac{L}{4}+1].
\label{cons1}
\end{equation}

$\mathbb{D}_{1,4}\cup\mathbb{D}_{2,4}\cup\mathbb{D}_{3,4}\cup\mathbb{D}_{4,4}\cup\mathbb{D}_{4,5}\cup\mathbb{D}_{4,6}\cup\mathbb{D}_{4,7}$ generates a consecutive range $\mathbb{R}_2$, as seen in (\ref{cons2}).
\begin{equation}
[\frac{L^2}{8}-\frac{L}{4}+1,-\frac{7L^2}{8} + (K-\frac{1}{4})L].
\label{cons2}
\end{equation}

$\mathbb{D}_{1,6}\cup\mathbb{D}_{2,5}\cup\mathbb{D}_{2,6}\cup\mathbb{D}_{3,5}\cup\mathbb{D}_{3,6}\cup\mathbb{D}_{3,7}$ generates a consecutive range $\mathbb{R}_3$, as given in (\ref{cons3}).
\begin{equation}
[-\frac{7L^2}{8} + (K-\frac{1}{4})L + 1,-\frac{{3{L^2}}}{4} + (K-\frac{1}{2})L + 3].
\label{cons3}
\end{equation}

Finally, it is proved that $\mathbb{C}_{\text{EMISC}}^{+}=\mathbb{R}_1\cup\mathbb{R}_2\cup\mathbb{R}_3 \subset \mathbb{D}_{\text{EMISC}}^{+}$. Moreover, $\mathbb{C}_{\text{EMISC}}^{+}$ is the maximum consecutive part of $\mathbb{D}_{\text{EMISC}}^{+}$, owing to a hole at the position of $-\frac{{3{L^2}}}{4} + (K-\frac{1}{2})L + 4$.

{The expressions of self-DCAs among ULSAs in the proposed EMISC SA are provided as}
\begin{flalign*}
&\mathbb{D}_{\text{1,1}}={\Big\{ 0,3 \Big\},}\\
&\mathbb{D}_{\text{2,2}}={\Big\{ 0,2,\dots,\frac{L}{2}-4 \Big\},}\\
&\mathbb{D}_{\text{3,3}}={\Big\{ 0,1,\dots,\frac{L}{4}-2)(\frac{L}{2}+1) \Big\},}\\
&\mathbb{D}_{\text{4,4}}={\Big\{(0,1,\dots,K-L-1)L \Big\},}\\
&\mathbb{D}_{\text{5,5}}={\Big\{(0,1,\dots,\frac{L}{4}-2)(\frac{L}{2}-1) \Big\},}\\
&\mathbb{D}_{\text{6,6}}={\Big\{ 0,1 \Big\},}\\
&\mathbb{D}_{\text{7,7}}={\Big\{ 0,2,\dots,\frac{L}{2}-4 \Big\}.}\\
\end{flalign*}
\begin{flalign*}
&{{\quad\text{The expressions of cross-DCAs are provided as}}}\\
&\mathbb{D}_{\text{1,2}}=\Big\{ 2+i+j \Big\},\\
&\mathbb{D}_{\text{1,3}}=\Big\{ \frac{L}{2}+i+j \Big\},\\
&\mathbb{D}_{\text{1,4}}=\Big\{ \frac{L^2}{8}+\frac{L}{4}-1+i+j \Big\},\\
&\mathbb{D}_{\text{1,5}}=\Big\{ -\frac{7L^2}{8}+(K+\frac{1}{4})L-1+i+j \Big\},\\
&\mathbb{D}_{\text{1,6}}=\Big\{ -\frac{3L^2}{4}+(K-\frac{1}{2})L-1+i+j \Big\},\\
&\mathbb{D}_{\text{1,7}}=\Big\{ -\frac{3L^2}{4}+(K-\frac{1}{2})L+3+i+j \Big\},\\
&\mathbb{D}_{\text{2,3}}=\Big\{ 2+i+j \Big\},\\
&\mathbb{D}_{\text{2,4}}=\Big\{ \frac{L^2}{8}-\frac{L}{4}+1+i+j \Big\},\\
&\mathbb{D}_{\text{2,5}}=\Big\{ -\frac{7L^2}{8}+(K-\frac{1}{4})L+1+i+j \Big\},\\
&\mathbb{D}_{\text{2,6}}=\Big\{ -\frac{3L^2}{4}+(K-1)L+1+i+j \Big\},\\
&\mathbb{D}_{\text{2,7}}=\Big\{ -\frac{3L^2}{4}+(K-1)L+5+i+j \Big\},\\
&\mathbb{D}_{\text{3,4}}=\Big\{ \frac{L}{2}+1+i+j \Big\},\\
&\mathbb{D}_{\text{3,5}}=\Big\{ -L^2+(K+\frac{1}{2})L+1+i+j \Big\},\\
&\mathbb{D}_{\text{3,6}}=\Big\{ -\frac{7L^2}{8}+(K-\frac{1}{4})L+1+i+j \Big\},\\
&\mathbb{D}_{\text{3,7}}=\Big\{ -\frac{7L^2}{8}+(K-\frac{1}{4})L+5+i+j \Big\},\\
&\mathbb{D}_{\text{4,5}}=\Big\{ L+i+j \Big\},\\
&\mathbb{D}_{\text{4,6}}=\Big\{ \frac{L^2}{8}+\frac{L}{4}+i+j \Big\},\\
&\mathbb{D}_{\text{4,7}}=\Big\{ \frac{L^2}{8}+\frac{L}{4}+4+i+j \Big\},\\
&\mathbb{D}_{\text{5,6}}=\Big\{ \frac{L}{2}-2+i+j \Big\},\\
&\mathbb{D}_{\text{5,7}}=\Big\{ \frac{L}{2}+2+i+j \Big\},\\
&\mathbb{D}_{\text{6,7}}=\Big\{ 3+i+j \Big\}.\\
& {{\text{where,}}}\\
&\mathbb{D}_{\text\it{m,n}} \hspace{0.5em} (m \leq n, \hspace{0.5em} m=1,2,\dots,6,7; \hspace{0.5em} n=1,2,\dots,6,7.)\\
& 1)\quad m = 1, \quad i =(0,3);\\
& 2)\quad m = 2, \quad i =(0,2,\dots,\frac{L}{2}-4);\\
& 3)\quad m = 3, \quad i =(0,1,\dots,\frac{L}{4}-2)(\frac{L}{2}+1);\\
& 4)\quad m = 4, \quad i =(0,1,\dots,K-L-1)(L);\\
& 5)\quad m = 5, \quad i =(0,1,\dots,\frac{L}{4}-2)(\frac{L}{2}-1);\\
& 6)\quad m = 6, \quad i =(0,1);\\
& 7)\quad m = 7, \quad i =(0,2,\dots,\frac{L}{2}-4).\\
&{{\text{Likewise, the variables $n$ and $j$ are presented like $m$ and $i$ above.}}}
\end{flalign*}

\subsection{uDOF Derivation}
Based on the proof in Section III. B, (\ref{consecutive_part1}) gives the expression of the consecutive part $\mathbb{C}_{\text{EMISC}}$
\begin{equation}
\mathbb{C}_{\text{EMISC}}=[\frac{{3{L^2}}}{4} - (K-\frac{1}{2})L - 3,-\frac{{3{L^2}}}{4} + (K-\frac{1}{2})L + 3].
\label{consecutive_part1}
\end{equation}

Then, the maximum uDOF of EMISC SA is provided by
\begin{equation}
{\text{uDOF}}_{\text{EMISC}}=-\frac{{3{L^2}}}{2} + (2K-1)L + 7.
\label{uDOF1}
\end{equation}

Interpolating (\ref{M}) into (\ref{uDOF1}), ${\text{uDOF}}_{\text{EMISC}}$ is given as
\begin{equation}
\begin{aligned}
{{\text{uDOF}}_{\text{EMISC}}=\left\{ {\begin{array}{*{20}{l}}
\frac{{{\rm{2}}{K^2}}}{3} - \frac{{2K}}{3} + 3,&K\% 6 = 3,4\\
\frac{{{\rm{2}}{K^2}}}{3} - \frac{{2K}}{3} + \frac{17}{3},&K\% 6 = 2,5,\\
\frac{{{\rm{2}}{K^2}}}{3} - \frac{{2K}}{3} + 7,&K\% 6 = 0,1
\end{array}} \right.}
\label{uDOF_EMISC}
\end{aligned}
\end{equation}

Comparatively, the uDOF of the IMISC SA is presented as
\begin{equation}
\begin{aligned}
{{\text{uDOF}}_{\text{IMISC}}=\left\{ {\begin{array}{*{20}{l}}
\frac{{{\rm{2}}{K^2}}}{3} - \frac{{2K}}{3} - 1,&K\% 6 = 3,4\\
\frac{{{\rm{2}}{K^2}}}{3} - \frac{{2K}}{3} + \frac{5}{3},&K\% 6 = 2,5.\\
\frac{{{\rm{2}}{K^2}}}{3} - \frac{{2K}}{3} + 3,&K\% 6 = 0,1
\end{array}} \right.}
\label{uDOF_IMISC}
\end{aligned}
\end{equation}

(\ref{uDOF_EMISC}) and (\ref{uDOF_IMISC}) show that {both EMISC and IMISC SAs} are able to utilize $K$ elements to generate $O(\frac{2K^2}{3})$ uDOFs, higher than most existing SAs. Meanwhile, it's found that the EMISC SA has four more uDOFs than the IMISC SA.

\subsection{Weight Function}
Considering that the first three weight values 
 {$w(i)(i=1,2,3)$} jointly dominate the MC matrix. Thus, {$w(i)(i=1,2,3)$} of the EMISC SA are provided by
\begin{equation}
\begin{aligned}
&w(1)=1,~ w(2)=2, \quad\quad\quad\hspace{0.8em}~ w(3)=2,&10\le K < 16,\\
&w(1)=1,~ w(2)=2\lfloor \frac{K-4}{6}\rfloor, ~w(3)=2,&K \ge 16.
\label{w1}
\end{aligned}
\end{equation}

Likewise, {$w(i)(i=1,2,3)$} of the IMISC SA are listed as
\begin{equation}
\begin{aligned}
&w(1)=2,~ w(2)=5, \quad\quad\quad\hspace{0.7em}~ w(3)=2,&10\le K < 16,\\
&w(1)=2, ~w(2)=2\lfloor \frac{K+2}{6}\rfloor,~ w(3)=1,&K \ge 16.
\label{w2}
\end{aligned}
\end{equation}

(\ref{w1}) and (\ref{w2}) confirm that the EMISC SA has a much lower MC than the IMISC SA. Thus, the CL of MISC-based SAs ranks in such an order: EMISC$<$IMISC$<$MISC.

\section{Numerical Simulations}\label{NE}
The proposed EMISC SA is compared with ANAI-2 [13], TSENA [14], {f-ENAMS-II [15], SNAs [16]-[18],} ePCA [19], ICNA [20], UF-4BL [21], [22], MISC [23] and IMISC [24] SAs. Besides, DOA estimation performance is analyzed via the spatial-smoothing MUSIC (SS-MUSIC), utilizing 1000 snapshots of data, and a coupling parameter ~$u_k=u_1e^{-j(k-1)\pi/8}/k, \, k=2,\dots,100$.
\subsection{uDOF and CL}
The EMISC SA is compared with other existing SAs in Fig. 2, from aspects of uDOF and CL.
\captionsetup[figure]{name={Fig.}, labelformat=simple, labelsep=period, singlelinecheck=off}
\begin{figure}[!t]
\centering
    \subfigure[uDOF.]{\includegraphics[width=0.48\columnwidth,height=4.0cm]{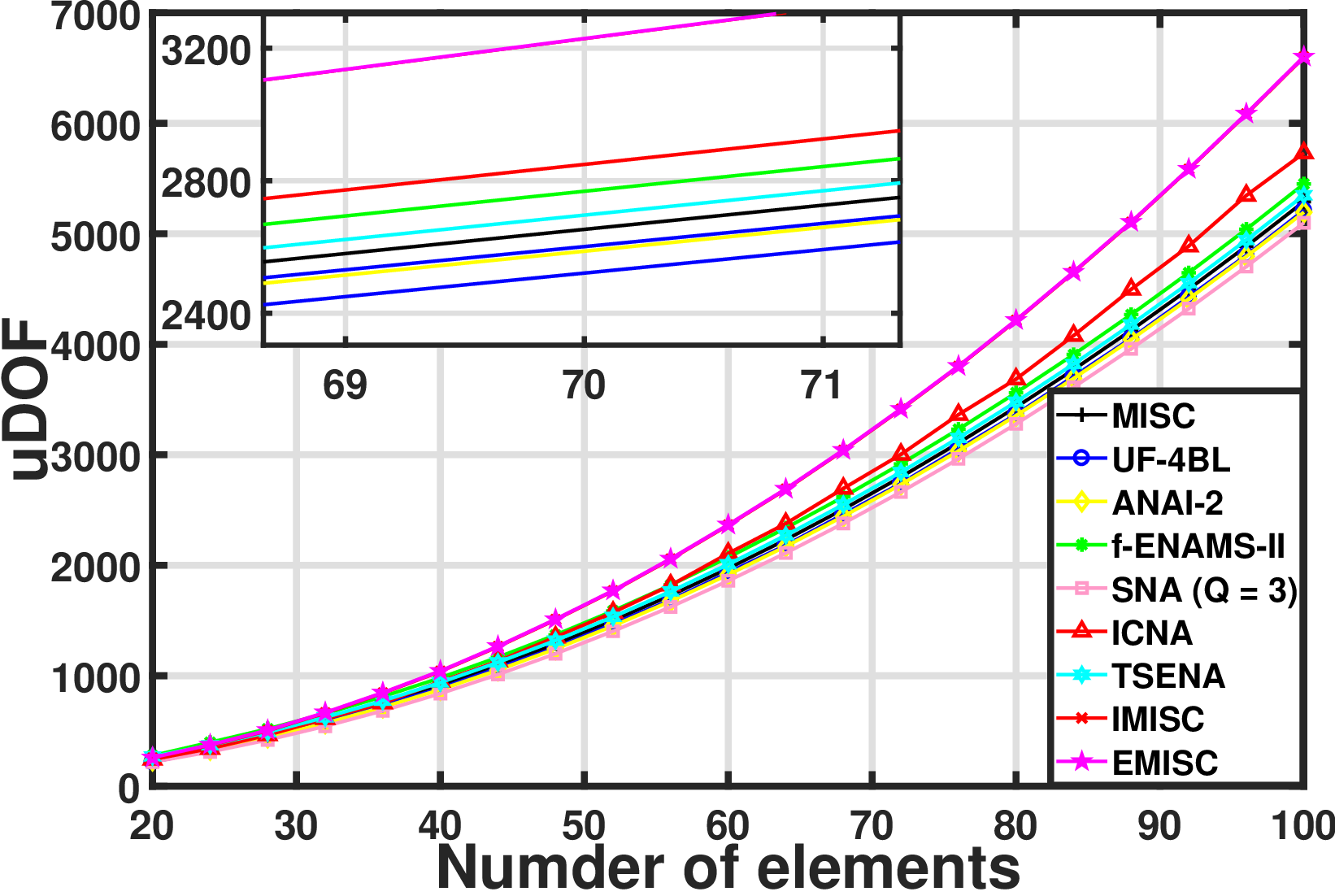} }
    \subfigure[CL.]{\includegraphics[width=0.48\columnwidth,height=4.0cm]{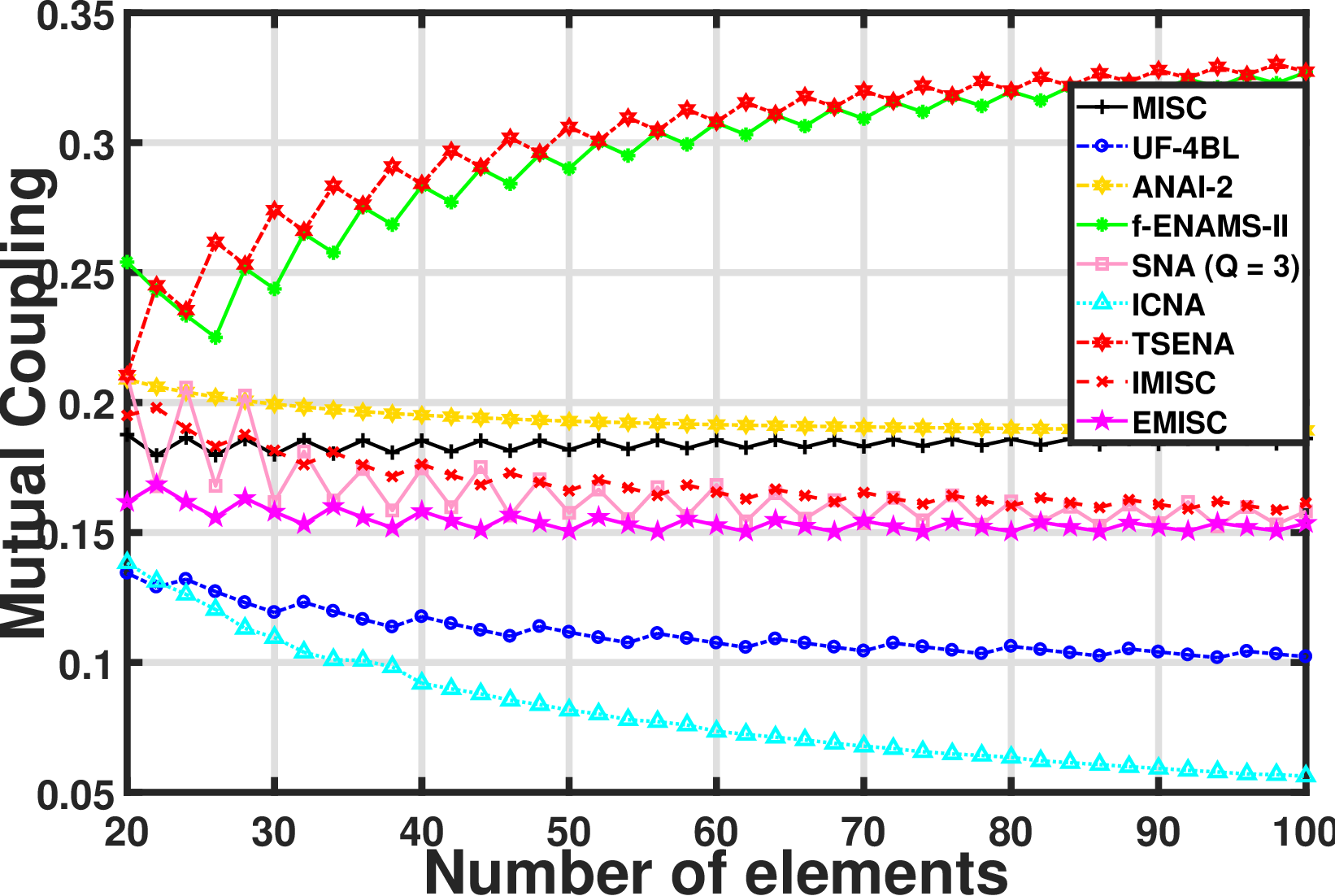} }
    \caption{uDOF and CL versus number of elements.}
    \label{uDOF-CL}
\end{figure}

Fig. 2(a) provides the curves that illustrate the effect of the element number on the uDOF. It is noted that the EMISC SA behaves best among all SAs. {As the number of elements increases}, the advantages of the EMISC SA gradually become prominent, owing to the $O(2K^2/3)$ uDOF of the EMISC SA, superior to the maximum uDOF $O(4K^2/7)$ of other existing SAs.

Fig. 2(b) reflects how the number of element affects the CL. The CL of the EMISC SA is lower than that of the IMISC SA, but still higher than that of ICNA and UF-4BL. In sum, Fig. 2 confirms that the EMISC SA outperforms other existing SAs, in terms of uDOF and CL.
\subsection{RMSE versus SNR and $|u_1|$}
Here, through 500 Monte Carlo experiments, Fig. 3 assesses the DOA estimation performance by RMSE versus SNR and $|u_1|$. 48 sources uniformly distributed among $[-60^{o} 60^{o}]$ are received by 36-element SAs. Under a fixed coupling parameter ~$u_1=0.3e^{j\pi/3}$, the RMSE versus SNR is illustrated in Fig. 3(a). For a fixed SNR=0 dB, the RMSE versus $|u_1|$ ranging from 0 to 0.5 is demonstrated in Fig. 3(b).
\captionsetup[figure]{name={Fig.}, labelformat=simple, labelsep=period, singlelinecheck=off}
\begin{figure}[!t]
\centering
    \subfigure[RMSE versus SNR.]{\includegraphics[width=0.48\columnwidth,height=4.3cm]{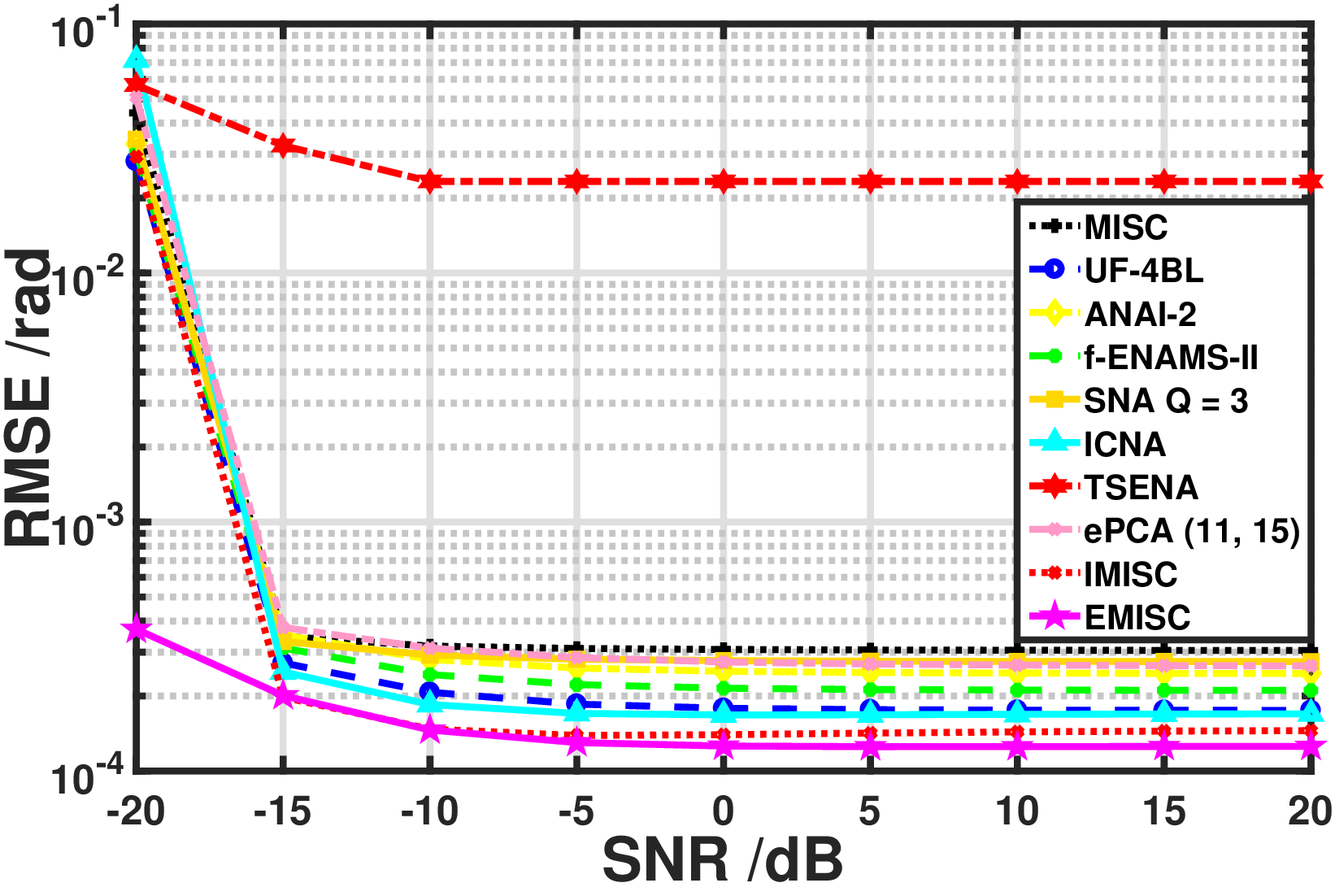} }
    \subfigure[RMSE versus $|u_1|$.]{\includegraphics[width=0.48\columnwidth,height=4.3cm]{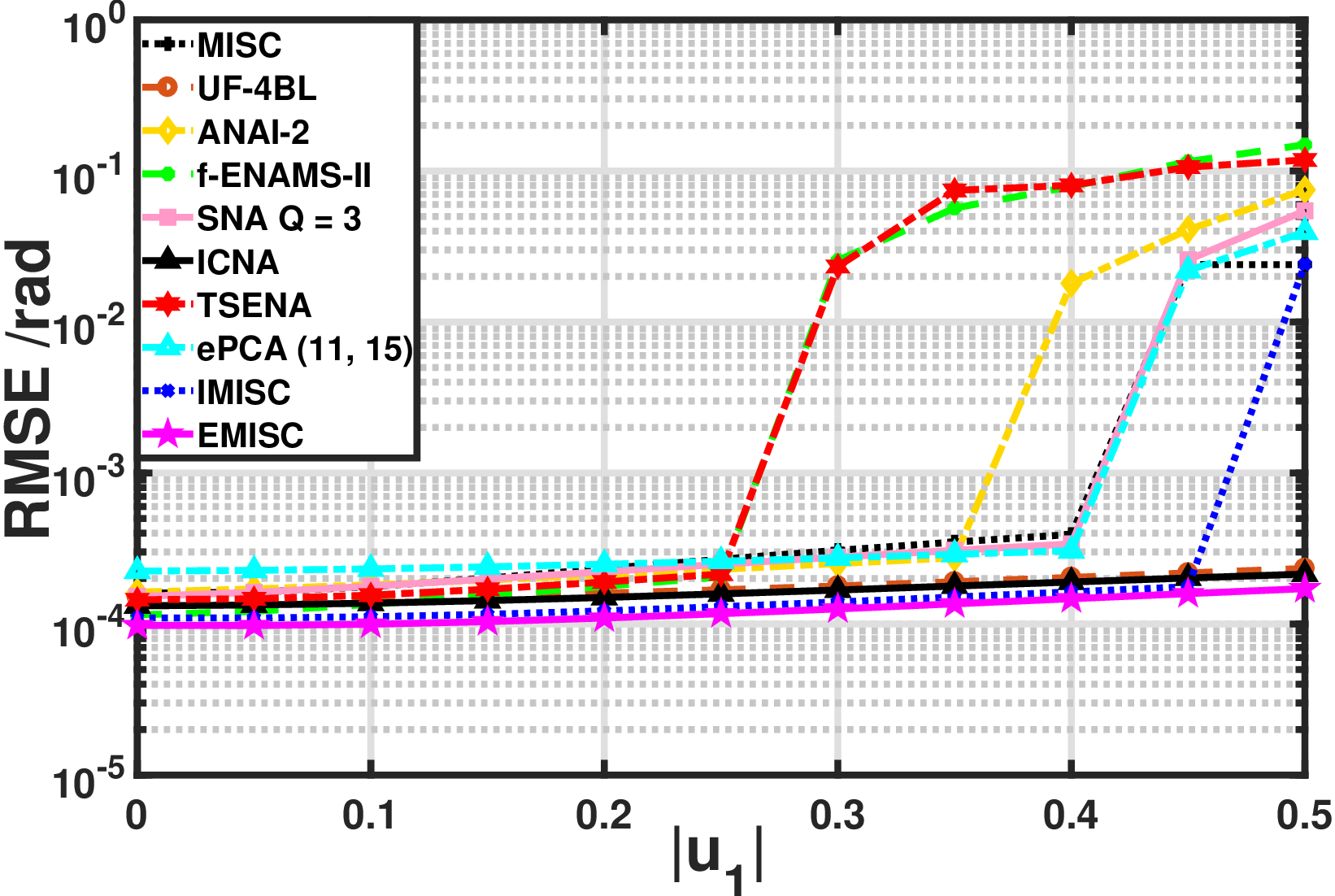} }
    \caption{RMSE performance versus SNR and $|u_1|$.}
    \label{SNR-$|u_1|$}
\end{figure}

To sum up, compared to all mentioned SAs above, Fig. 3 verifies that the proposed EMISC SA possesses the lowest RMSE, from aspects of SNR and $|u_1|$.

\section{Data Processing}\label{Data Processing}
Here, DOA estimation performance of the proposed EMISC SA is verified by real data collected in an underwater target detection trial on Sep. 24-25th, 2020. A 15-element receiver SA was deployed at the seabed. The observed angle is measured from the broadside of the receiver SA, within a range of [-90\textdegree,90\textdegree]. Occasionally, there appeared some operating fishing boats. The received data was composed of ninety 20s-packages and processed at a frequency band of [50Hz,120Hz].

Fig. 4 gives the bearing-time records (BTRs). It is observed that there exists a weak target (-3\textdegree) and two neighboring targets (63\textdegree, 68\textdegree). The weak target gets detected in the BTRs of all SAs in Fig. 4. Due to the variance of uDOF, two neighboring targets cannot be distinguished in Figs. 4(a), (b) and (c), while it works in Figs. 4(d), (e) and (f). Compared to Figs. 4(e) and (f), the bearing resolution of two neighboring targets appears blurry in Fig. 4(d). {Considering no MC in the underwater acoustic arrays,} the EMISC SA basically performs as well as the IMISC SA, owing to the approximately same uDOFs.
\captionsetup[figure]{name={Fig.}, labelformat=simple, labelsep=period, singlelinecheck=off}
\begin{figure}[!t]
\centering
    \subfigure[ANAI2.]{\includegraphics[width=0.46\columnwidth,height=2.8cm]{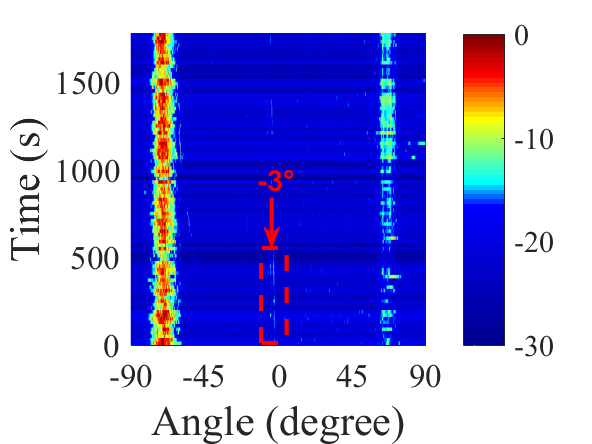} }
    \subfigure[UF-4BL.]{\includegraphics[width=0.46\columnwidth,height=2.8cm]{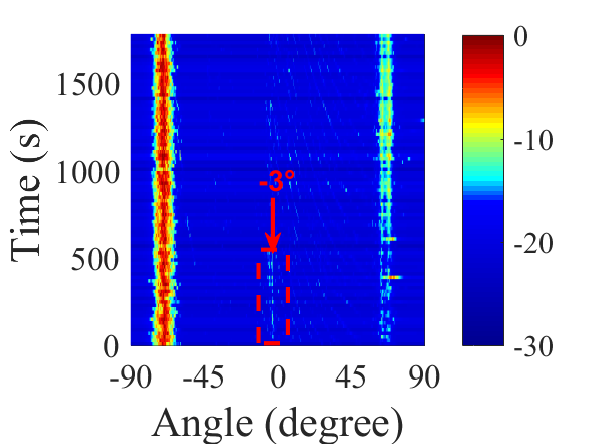} }
    \subfigure[ICNA.]{\includegraphics[width=0.46\columnwidth,height=2.8cm]{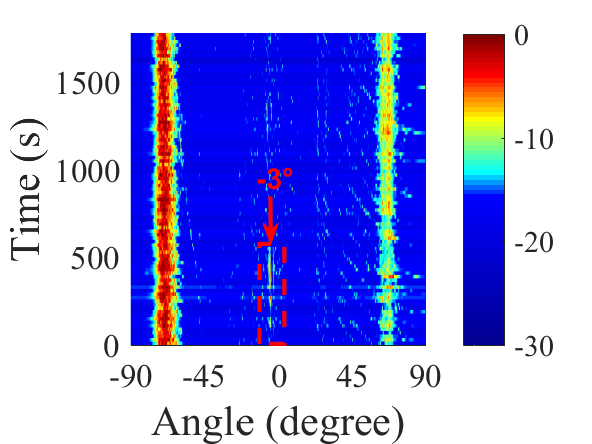} }
    \subfigure[MISC.]{\includegraphics[width=0.46\columnwidth,height=2.8cm]{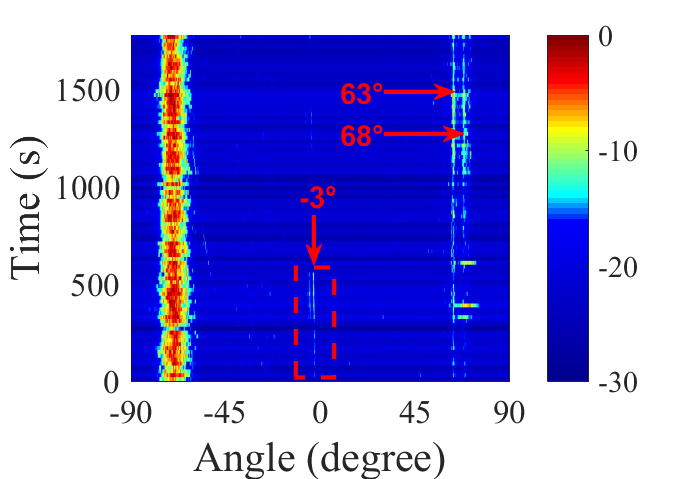} }
    \subfigure[IMISC.]{\includegraphics[width=0.46\columnwidth,height=2.8cm]{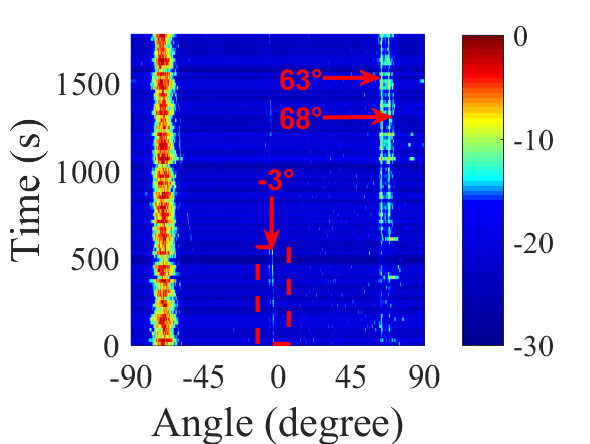} }
    \subfigure[EMISC.]{\includegraphics[width=0.46\columnwidth,height=2.8cm]{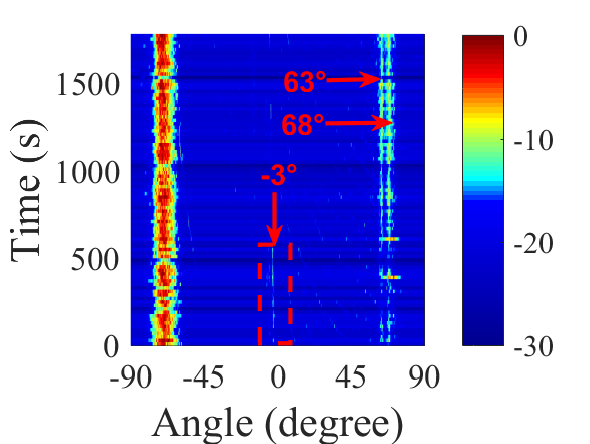} }
    \caption{The bearing-time records (BTRs).}
    \label{BTRs}
\end{figure}

\section{Conclusion}\label{Conclusion}
In this letter, an enhanced MISC-based (EMISC) sparse array (SA) with high uDOFs and low MC effect is proposed. The EMISC SA is designed via an IES set, determined by the maximum IES and the number of element. The uDOF and weight-function of the EMISC SA are derived in theory. Compared to the IMISC SA, the EMISC SA further increases the uDOFs and reduces the MC. For DOA estimation, the EMISC SA has a much lower RMSE than other existing SAs. Finally, the EMISC SA is verified by real data.

%\begin{appendices}
%\end{appendices}
%%%%%%%%%%%%%%%%%%%%%%%%%%%%%%%%%%%%%%%%%%%%%%%
%%%%%%%%%%%%%%%%%%%%%%%%%%%%%%%%%%%%%%%%%%%%%%%

\end{document}